\documentclass{article}

\usepackage{arxiv}

\usepackage[utf8]{inputenc} % allow utf-8 input
\usepackage[T1]{fontenc}    % use 8-bit T1 fonts
\usepackage{hyperref}       % hyperlinks
\usepackage{url}            % simple URL typesetting
\usepackage{booktabs}       % professional-quality tables
\usepackage{amsfonts}       % blackboard math symbols
\usepackage{nicefrac}       % compact symbols for 1/2, etc.
\usepackage{microtype}      % microtypography
\usepackage[normalem]{ulem}
\useunder{\uline}{\ul}{}
\usepackage{graphicx}
\graphicspath{ {./images/} }

\title{AI-Assisted Verification of Biometric Data Collection}

\author{
  Ryan M. Lindsey\thanks{https://github.com/rynsy} \\
  University of Kentucky
}

\begin{document}
\maketitle

%\begin{abstract}

%\end{abstract}

% keywords can be removed

\section{Abstract}
Recognizing actions from a video feed is a challenging task to automate, especially so on older hardware. There are two aims for this project: one is to recognize an action from the front-facing camera on an Android phone, the other is to support as many phones and Android versions as possible. This limits us to using models that are small enough to run on mobile phones with and without GPUs, and only using the camera feed to recognize the action. In this paper we compare performance of the YOLO architecture across devices (with and without dedicated GPUs) using models trained on a custom dataset. We also discuss limitations in recognizing faces and actions from video on limited hardware.

\section{Motivation}
\label{sec:problem}

\begin{figure}
    \centering
    \includegraphics[width=6cm]{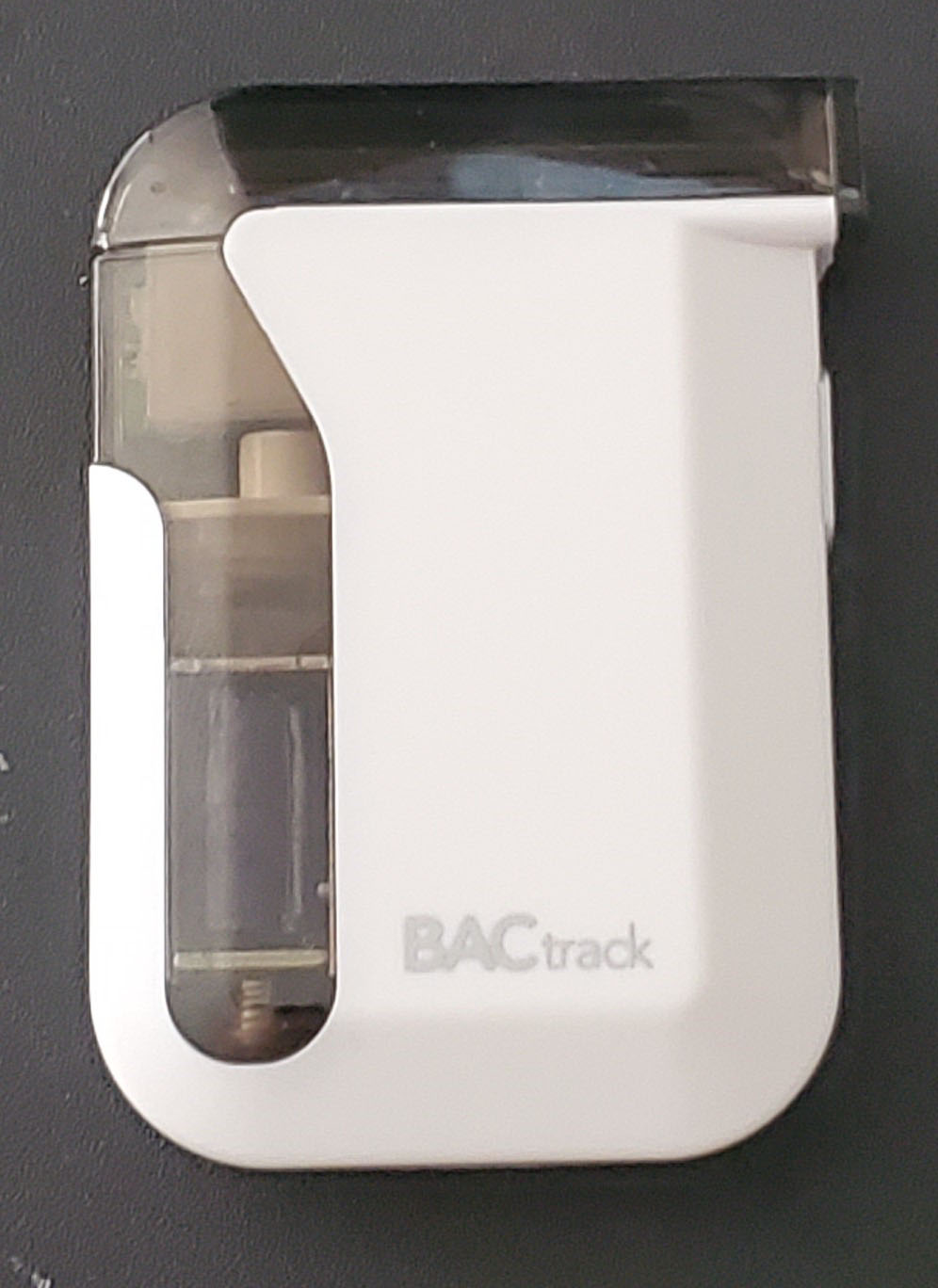}
    \caption{BACtrack Breathalyzer}
    \label{fig:breathalyzer}
\end{figure}

There is an Android app that was created to help its users quit drinking alcohol by providing financial incentives. Users are given a BACtrack breathalyzer (see Figure \ref{fig:breathalyzer}) that connects to their phone over Bluetooth. The user submits BAC readings to the app on a regular basis, and they are rewarded for reaching milestones in their treatment (1 week sober, 1 month sober, 3 months sober, etc). Given the incentives, the user could game the system by having a sober friend submit samples for them and still reap the rewards of maintaining sobriety. This gives us motivation to secure this process as best we can, but we're limited in options to do so.
\par
Ideally, this application could run on as many Android devices as possible. As of this writing there are built-in facilities for doing facial recognition on newer Android devices, but not all participants will have access to the latest phones or Android software. These older phones are unlikely to have dedicated GPUs as well, so our model will have to be able to run on the CPU and perform decently well. Over time more Android devices will have dedicated GPUs and built-in facial recognition, but for the users without such phones we'll have to create something that can accomplish the same goals. We can create a model that can be used in both cases:

\begin{enumerate}
    \item Prompt user to submit sample
    \item Turn on front-facing camera and capture an image
    \item Using a lightweight model, verify that the breathalyzer and a face are in-frame. 
    \begin{enumerate}
        \item If the phone has facial recognition, use the Android Biometric library to verify that the face in frame is the owner of the phone (the participant in the program)
        \item If not, send a collection of frames to a server to verify that the person submitting the sample is the correct participant. 
    \end{enumerate}
\end{enumerate}

In the roadmap towards solving this problem we need a few things: 
\begin{itemize}
    \item A network architecture that is performant enough to classify objects in a video feed.
    \item To quickly identify this particular model of breathalyzer and a face in a video feed (this work)
    \item To convert this model to a format (PyTorch, Tensorflow Lite) that can run on an Android device.
\end{itemize}

\par
There are many models that can be used to recognize faces in an image, for this project it would save time to re-train those models to recognize an additional label (the breathalyzer). The performance of the model/architecture and how well the model can be converted and run in a mobile context are also big concerns for this project. After comparing options, the YOLO architecture\cite{yolo} seemed most appropriate. A custom dataset of the BACtrack breathalyzer in multiple poses and contexts was built for this project, consisting of about 800 images, augmented using the Roboflow\cite{roboflow} platform.

\begin{figure}
    \centering
    \includegraphics[width=10cm]{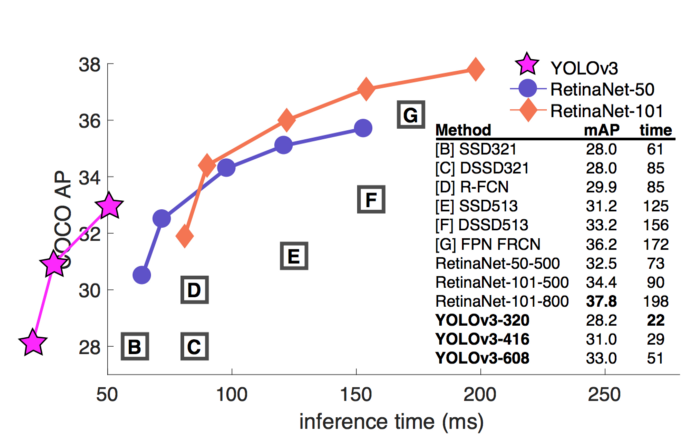}
    \caption{Performance comparison of YOLO-v3 to other object detection models on the COCO dataset\cite{yolo}}
    \label{fig:yolo_performance}
\end{figure}

\section{Experiment}

The dataset was built by taking pictures of the BACtrack breathalyzer in a variety of poses and contexts with around 800 images in the dataset. The images were labeled using LabelImg\cite{labelimg} and exported to Pascal/VOC format. Custom scripts were written to convert these to a format that is compatible with YOLOv3\cite{yolo} and to perform the testing/training split of the dataset. Only the last few layers were re-trained, which required a custom configuration to freeze the layers that did not require retraining\cite{RyanDarknet}. The size of the model proved to be an issue, so the training had to occur on one of the University of Kentucky's high performance computing (HPC) clusters.
\par
Training the model requires a lot of preparation: downloading the dataset, downloading dependencies and configuring them, creating the test/train split, updating the configuration of Darknet. This process was scripted and built into a custom Docker container. The HPC cluster runs jobs using Singularity\cite{singularity} and SLURM jobs, so the container was converted using SingularityWare's Docker-to-Singularity conversion tool. At this stage we have a Singularity container to handle setting up the project, training the model, and sending the final model to long-term storage.
\par
The resulting model performs reasonably well on a standard desktop, around 20FPS when run on a webcams video feed. This model is too large and slow to load on an Android device though, the weights of the model take 250Mb to store. This process was re-run using configurations for the YOLOv3-tiny\cite{yolo}, which gives a 35Mb file with similar accuracy that can be compressed to run on an Android device.
\par

\section{Results}
\label{sec:results}

\begin{table}[h]
\centering
\begin{tabular}{|c|c|c|c|c|c|c|}
\hline
{\ul \textbf{Model}} & {\ul \textbf{Quadro P6000}} & {\ul \textbf{TX2 (G)}} & {\ul \textbf{Nano (G)}} & {\ul \textbf{Intel i7-4790k}} & {\ul \textbf{TX2}} & {\ul \textbf{Nano}}  \\ \hline
YOLO-v3 Breath & 28ms & 323ms & 738ms & 1000ms & 2502ms & 3407ms \\ \hline
YOLO-v3-small Breath & 4ms & 38ms & 91ms & 167ms & 351ms & 482ms \\ \hline

\end{tabular}
\caption{Comparison of inference times across devices}
\label{tab:inference}
\end{table}

\section{Future Work}
\label{sec:future}
In the short time since this original experiment there have been two major revisions of YOLO. The current version (YOLOv5\cite{yolov5}) has performance gains over YOLOv3 and convenient functions for converting the weights to a variety of formats (TFLite, ONNX, TensorRT). The most obvious next step is to train a new model using this framework and gather inference times across devices (including Android). This model could also be used to build a prototype verification system that could work on devices with and without built-in facial recognition hardware and software. For devices with the special hardware, use this model to verify presence of the breathalyzer, and built-in facilities to verify that the face in frame is the owner of the device. The process is the same for devices without facial recognition, but verification that the correct person is submitting a sample would be delegated to a service running in the cloud.
\par

\bibliographystyle{unsrt}  
\bibliography{references}

\end{document}